\documentclass[10pt,twocolumn,letterpaper]{article}

\usepackage{cvpr}
\usepackage{times}
\usepackage{epsfig}
\usepackage{graphicx}
\usepackage{amsmath}
\usepackage{amssymb}
\usepackage{subfig}

\usepackage[pagebackref=true,breaklinks=true,letterpaper=true,colorlinks,bookmarks=false]{hyperref}

 \cvprfinalcopy % *** Uncomment this line for the final submission

\def\cvprPaperID{1756} % *** Enter the CVPR Paper ID here

\ifcvprfinal\fi
\begin{document}

%%%%%%%%% TITLE
\title{A Hierarchical Deep Temporal Model for Group Activity Recognition}

\author{Mostafa S. Ibrahim\thanks{Equal Contribution} , Srikanth Muralidharan\footnotemark[1] , Zhiwei Deng, Arash Vahdat, Greg Mori\\
School of Computing Science, Simon Fraser University, Burnaby, Canada \\
{\tt\small \{msibrahi, smuralid, zhiweid, avahdat\}@sfu.ca, mori@cs.sfu.ca}}

\maketitle
%\thispagestyle{empty}
%%%%%%%%% ABSTRACT
\begin{abstract}
   In group activity recognition, the temporal dynamics of the whole activity can be inferred based on the dynamics of the individual people representing the activity. We build a deep model to capture these dynamics based on LSTM (long short-term memory) models. To make use of these observations, we present a 2-stage deep temporal model for the group activity recognition problem. In our model, a LSTM model is designed to represent action dynamics of individual people in a sequence and another LSTM model is designed to aggregate person-level information for whole activity understanding.
  We evaluate our model over two datasets: the Collective Activity Dataset and a new volleyball dataset. Experimental results demonstrate that our proposed model improves group activity recognition performance  compared to baseline methods.
\end{abstract}

%%%%%%%%% BODY TEXT
\section{Introduction}

\begin{figure}[t]
  \includegraphics[width=1.0\linewidth]{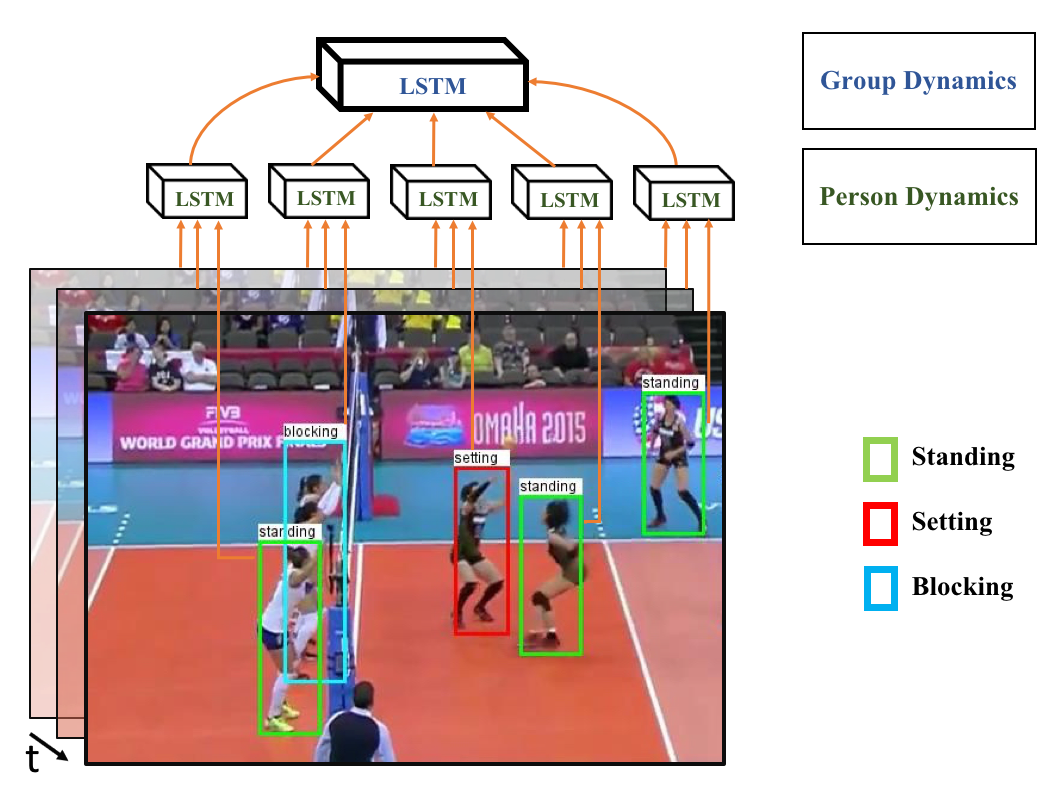}
    \caption{Group activity recognition via a hierarchical model. Each person in a scene is modeled using a temporal model that captures his/her dynamics, these models are integrated into a higher-level model that captures scene-level activity.}
  \label{fig_illus}
  \end{figure}

  What are the people in Figure~\ref{fig_illus} doing?  This question can be answered at numerous levels of detail -- in this paper we focus on the group activity, a high-level answer such as ``team spiking acivity''.  We develop a novel hierarchical deep model for group activity recognition.

  A key cue for group activity recognition is the spatio-temporal relations among the people in the scene.  Determining where individual people are in a scene, analyzing their image appearance, and aggregating these features and their relations can discern which group activity is present.  A volume of research has explored models for this type of reasoning~\cite{choi2009they,lan2012social,ramanathan2013social,amer2014hirf}.  However, these approaches have focused on probabilistic or discriminative models built upon hand-crafted features.  Since they rely on shallow hand crafted feature representations, they are limited by their representational abilities to model a complex learning task. Deep representations have overcome this limitation and yielded state of the art results in several computer vision benchmarks~\cite{krizhevsky2012imagenet,simonyan2014very,karpathy2014large}.
  
  A naive approach to group activity recognition with a deep model would be to simply treat an image as an holistic input.  One could train a model to classify this image according to the group activity taking place. However, it isn't clear if this will work given the redundancy in the training data: with a dataset of volleyball videos, frames will be dominated by features of volleyball courts. The differences between the different classes of group activities are about spatio-temporal relations between people, beyond just global appearance.  Forcing a deep model to learn invariance to translation, to focus on the relations between people, presents a significant challenge to the learning algorithm.  Similar challenges exist in the object recognition literature, and research often focuses on designing pooling operators for deep networks (e.g.~\cite{Szegedy15}) that enable the network to learn effective classifiers.
  
  Group activity recognition presents a similar challenge -- appropriate networks need to be designed that allow the learning algorithm to focus on differentiating higher-level classes of activities.  Hence, we develop a novel hierarchical deep temporal model that reasons over individual people.  Given a set of detected and tracked people, we run temporal deep networks (LSTMs) to analyze each individual person.  These LSTMs are aggregated over the people in a scene into a higher level deep temporal model.  This allows the deep model to learn the relations between the people (and their appearances) that contribute to recognizing a particular group activity.
  
  The main contribution of this paper is the proposal of a novel deep architecture that models group activities in a principled structured temporal framework. Our 2-stage approach models individual person activities in its first stage, and then combines person level information to represent group activities. The model's temporal representation is based on the long short-term memory (LSTM): recurrent neural networks such as these have recently demonstrated successful results in sequential tasks such as image captioning \cite{donahue2014long} and speech recognition
  \cite{graves2014towards}. Through the model structure, we aim at constructing a representation that leverages the discriminative information in the hierarchical structure between individual person actions and group activities. The model can be used in general group activity applications such as video surveillance, sport analytics, and video search and retrieval.
  
  To cater the needs of our problem, we also propose a new volleyball dataset that offers person detections, and both the person action label, as well as the group activity label. The camera view of the selected sports videos allows us to track the players in the scene. Experimentally, the model is effective in recognizing the overall team activity based on recognizing and integrating player actions.

  This paper is organized as follows. In Section~\ref{sec:related_work}, we provide a brief overview of the literature related to activity recognition. In Section~\ref{sec:model}, we elaborate details of the proposed group activity recognition model. In Section~\ref{sec:experiments}, we tabulate the performance of approach, and end in Section~\ref{sec:conclusion} with a conclusion of this work.

\section{Related Work}
\label{sec:related_work}
  Human activity recognition is an active area of research, with many existing algorithms. Surveys by Weinland et al.~\cite{weinland2011survey} and Poppe~\cite{poppe2010survey} explore the vast literature in activity recognition. Here, we will focus on the group activity recognition problem and recent related advances in deep learning.

\textbf{Group Activity Recognition:}
  Group activity recognition has attracted a large body of work recently. Most previous work has used hand-crafted features fed to structured models that represent information between individuals in space and/or time domains. Lan et al.~\cite{LanWYRM12} proposed an adaptive latent structure learning that represents hierarchical relationships ranging from lower person-level information to higher group-level interactions. Lan et al.~\cite{LanSM12} and Ramanathan et al.~\cite{ramanathan2013social} explore the idea of social roles, the expected behaviour of an individual person in the context of group, in fully supervised and weakly supervised frameworks respectively. Choi and Savarese~\cite{choiS12} have unified tracking multiple people, recognizing individual actions, interactions and collective activities in a joint framework.  In other work~\cite{choi2011learning}, a random forest structure is used to sample discriminative spatio-temporal regions from input video fed to 3D Markov random field to localize collective activities in a scene. Shu et al.~\cite{ShuXRTZ15} detect group activities from aerial video using an AND-OR graph formalism.   The above-mentioned methods use shallow hand crafted features, and typically adopt a linear model that suffers from representational limitations. 
  
\textbf{Sport Video Analysis:} Previous work has extended group activity recognition to team activity recognition in sport footage.  Seminal work in this vein includes Intille and Bobick~\cite{IntilleB01}, who examined stochastic representations of American football plays.
Siddiquie et al.~\cite{siddiquie2009recognizing} proposed sparse multiple kernel learning to select features incorporated in a spatio-temporal pyramid.
Morariu et al.~\cite{morariu11eventstructure} track players, infer part locations, and reason about temporal structure in 1-on-1 basketball games.
Swears et al.~\cite{SwearsHJB14} used the Granger Causality statistic to automatically constrain the temporal links of a Dynamic Bayesian Network (DBN) for handball videos. Direkoglu and O'Connor \cite{direkoǧluO12} solved a particular Poisson equation to generate a holistic player location representation. Kwak et al.~\cite{KwakHH13} optimize based on a rule-based depiction of interactions between people.
  
  \textbf{Deep Learning:} Deep Convolutional Neural Networks (CNNs) have shown impressive performance by unifying feature and classifier learning and the availability of large labeled datasets. Successes have been demonstrated on a variety of computer vision tasks including image classification~\cite{krizhevsky2012imagenet, simonyan2014very} and action recognition~\cite{simonyan2014two, karpathy2014large}. More flexible recurrent neural network (RNN) based models are used for handling variable length space-time inputs. Specifically, LSTM~\cite{hochreiter1997long} models are popular among RNN models due to the tractable learning framework that they offer when it comes to deep representations. These LSTM models have been applied to a variety of tasks \cite{donahue2014long,graves2014towards,ng2015beyond,venugopalan2014translating}.  For instance, in Donahue et al.~\cite{donahue2014long}, the so-called Long term Recurrent Convolutional network, formed by stacking an LSTM on top of pre-trained CNNs, is proposed for handling sequential tasks such as activity recognition, image description, and video description. In Karpathy et al.~\cite{karpathy2014deep}, structured objectives are used to align CNNs over image regions and bi-directional RNNs over sentences. A deep multi-modal RNN architecture is used for generating image descriptions using the deduced alignments.
  
  In this work, we aim at building a hierarchical structured model that incorporates a deep LSTM framework to recognize individual actions and group activities. Previous work in the area of deep structured learning includes Tompson et al.~\cite{NIPS2014_5573} for pose estimation, and Zheng et al.~\cite{Zheng15} and Schwing et al.~\cite{schwing2015fully} for semantic image segmentation. In Deng et al.~\cite{DengZCLMRM15} a similar framework is used for group activity recognition, where a neural network-based hierarchical graphical model refines person action labels and learns to predict the group activity simultaneously. While these methods use neural network-based graphical representations, in our current approach, we leverage LSTM-based temporal modelling to learn discriminative information from time varying sports activity data. In \cite{yeung2015every}, a new dataset is introduced that contains dense multiple labels per frame for underlying action, and a novel Multi-LSTM is used to model the temporal relations between labels present in the dataset. 
  
  \textbf{Datasets:}
    Popular datasets for activity recognition include the  Sports-1M dataset~\cite{karpathy2014deep}, UCF 101 database \cite{soomro2012ucf101}, and the HMDB movie database \cite{kuehne2011hmdb}. These datasets started to shift the focus to unconstrained Internet videos that contain more intra-class variation, compared to a constrained dataset. While these datasets continue to focus on individual human actions, in our work we focus on recognizing more complex group activities in sport videos. Choi et al.~\cite{choi2009they} introduced the Collective Activity Dataset consisting of real world pedestrian sequences where the task is to find the high level group activity. In this paper, we experiment with this dataset, but also introduce a new dataset for group activity recognition in sport footage which is annotated with player pose, location, and group activities to encourage similar research in the sport domain.

\section{Proposed Approach}
\label{sec:model}

Our goal in this paper is to recognize activities performed by a group of people in a video sequence.  The input to our method is a set of tracklets of the people in a scene.
The group of people in the scene could range from players in a sports video to pedestrians in a surveillance video. In this paper we consider three cues that can aid in determining what a group of people is doing:

\begin{itemize}
\item \textbf{Person-level actions} collectively define a group activity. Person action recognition is a first step toward recognizing group activities.

\item \textbf{Temporal dynamics of a person's action} is higher-order information that can serve as a strong signal for group activity. Knowing how each person's action is changing over time can be used to infer the  group's activity. 

% \item \textbf{Spatial context} can be represented as what most people in the scene are doing at the current time. MostafaComment: We use fc7 over person NOT scene.

\item \textbf{Temporal evolution of group activity} represents how a group's activity is evolving over time. For example, in a volleyball game a team may move from defence phase to pass and then attack. 

\end{itemize}

%Popular approaches that address activity recognition use hand crafted feature representation, and a linear model that has limited representational capability \cite{wang2011action} \cite{schuldt2004recognizing}. In the context of group activity recognition, it is shown that representing a group activity using a hierarchical models yield better performance than shallow representations \cite{LanWYRM12, LanSM12, ramanathan2013social}. \\

Many classic approaches to the group activity recognition problem have modeled these elements in a form of structured prediction based on hand crafted features \cite{wang2011action, schuldt2004recognizing, LanWYRM12, LanSM12, ramanathan2013social}. Inspired by the success of deep learning based solutions, in this paper, a novel hierarchical deep learning based model is proposed that is potentially capable of learning low-level image features, person-level actions, their temporal relations, and temporal group dynamics in a unified end-to-end framework.

Given the sequential nature of group activity analysis, our proposed model is based on a Recurrent Neural Network (RNN) architecture. RNNs consist of non-linear units with internal states that can learn dynamic temporal behavior from a sequential input with arbitrary length. Therefore, they overcome the limitation of CNNs that expect constant length input. This makes them widely applicable to video analysis tasks such as activity recognition. 

Our model is inspired by the success of hierarchical models. Here, we aim to mimic a similar intuition using recurrent networks. We propose a deep model by stacking several layers of RNN-type structures to model a large range of low-level to high-level dynamics defined on top of people and entire groups. We describe the use of these RNN structures for individual and group activity recognition next.

\subsection{Temporal Model of Individual Action} \label{sec:lstm}

Given tracklets of each person in a scene, we use long short-term memory (LSTM) models to represent temporally the action of each individual person. Such temporal information is complementary to spatial features and is critical for performance. LSTMs, originally proposed by Hochreiter and Schmidhuber \cite{hochreiter1997long}, have been used successfully for many sequential problems in computer vision. Each LSTM unit consists of several cells with memory that stores information for a short temporal interval. The memory content of a LSTM makes it suitable for modeling complex temporal relationships that may span a long range.

The content of the memory cell is regulated by several gating units that control the flow of information in and out of the cells. The control they offer also helps in avoiding spurious gradient updates that can typically happen in training RNNs when the length of a temporal input is large. This property enables us to stack a large number of such layers in order to learn complex dynamics present in the input in different ranges. 

We use a deep Convolutional Neural Network (CNN) to extract features from the bounding box around the person in each time step on a person trajectory. The output of the CNN, represented by $x_t$, can be considered as a complex image-based feature describing the spatial region around a person. Assuming $x_t$ as the input of an LSTM cell at time $t$, the cell activition can be formulated as :
  \begin{flalign}
  i_{t} &= \sigma(W_{xi}x_{t} + W_{hi}h_{t - 1} + b_{i}) \\
  f_{t} &= \sigma(W_{xf}x_{t} + W_{hf}h_{t - 1} + b_{f}) \\
  o_{t} &= \sigma(W_{xo}x_{t} + W_{ho}h_{t - 1} + b_{o}) \\ 
  g_{t} &= \phi(W_{xc}x_{t} + W_{hc}h_{t - 1} + b_{c}) \\
  c_{t} &= f_t \odot c_{t - 1} + i_t \odot g_t \\ 
  h_{t} &= o_t \odot \phi(c_{t})
  \end{flalign}
  
  Here, $\sigma$ stands for a sigmoid function, and $\phi$ stands for the tanh function. $x_t$ is the input, $h_t \in R^N$ is the hidden state with N hidden units, $c_t \in R^N$ is the memory cell, $i_t \in R^N$, $f_t \in R^N$, $o_t \in R^N$, and,  $g_t \in R^N$ are input gate, forget gate, output gate, and input modulation gate at time $t$ respectively. $\odot$ represents element-wise multiplication. 
  
  %[[ Is the below correct? ]]
  %I think what he meant was that we use these equations precisely for learning person level dynamics. Right?
  
  When modeling individual actions, the hidden state $h_t$ could be used to model the action a person is performing at time $t$. Note that the cell output is evolving over time based on the past memory content. Due to the deployment of gates on the information flow, the hidden state will be formed based on a short-range memory of the person's past behaviour.
  Therefore, we can simply pass the output of the LSTM cell at each time to a softmax classification layer\footnote{More precisely, a fully connected layer fed to softmax loss layer.} to predict individual person-level action for each tracklet.
  
  The LSTM layer on top of person trajectories forms the first stage of our hierarchical model. This stage is designed to model \textbf{person-level actions and their temporal evolution}. Our training proceeds in a stage-wise fashion, first training to predict person level actions, and then pasing the hidden states of the LSTM layer to the second stage for group activity recognition, as discussed in the next section.
  
  %Our model learns dynamic representations right from primitive actions to group activities using LSTMs. Through this approach, we show that LSTM's can be used simultaneously for feature learning, and also for constructing a discriminative representation of group activities. We next describe the pre-processing steps that we performed before conducting the experiment, and then follow it up with the description of our model.

  %Concretely, a semantic representation that captures different levels of interaction
 % (e.g. person-person interactions) distinctly is likely to be more successful than a model that does not possess such semantic structuring

\subsection{Hierarchical Model for Group Activity Recognition}

\begin{figure*}[t]
  \begin{center}
  \includegraphics[trim=0 0 0 0,clip, width=0.9\linewidth]{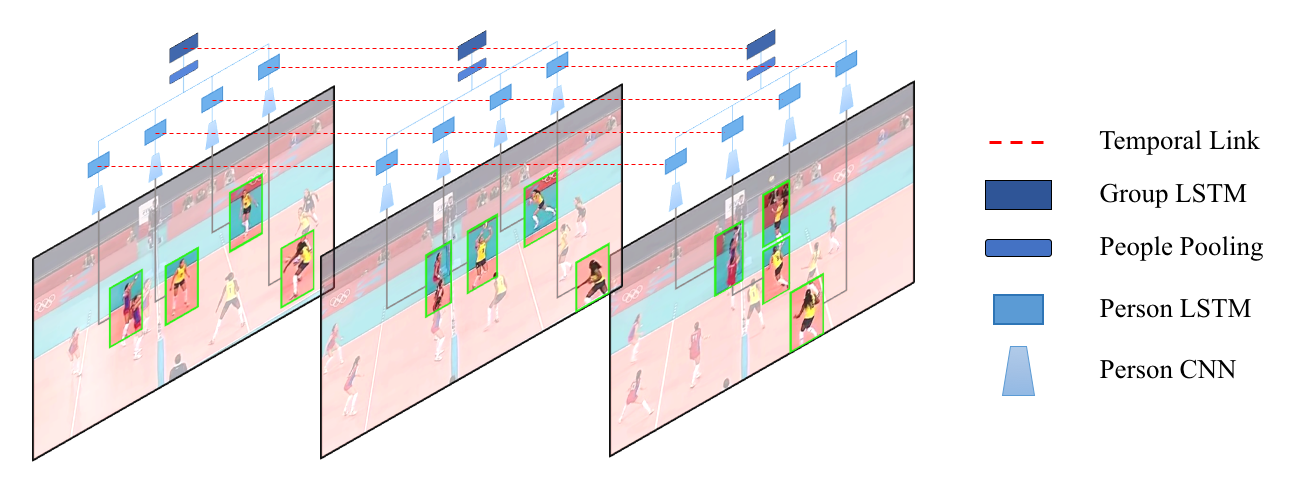}
  \end{center}
    \caption{Our two-stage model for a volleyball match. Given tracklets of K-players, we feed each tracklet in a CNN, followed by a person LSTM layer to represent each player's action. We then pool over all people's temporal features in the scene. The output of the pooling layer is feed to the second LSTM network to identify the whole teams activity.}
  \label{fig_two_stages_model}
  \end{figure*}

%The goal of group activity recognition is to assign a video clip to one of K activity classes. Given a video clip, we detect and track individuals in the scene. 

%We propose a hierarchical LSTM model for group activity recognition as shown in Figure~\ref{fig_two_stages_model}. The first level of our LSTM model is designed to model \textbf{person-level actions and their temporal evolution}. Given a person tracklet, we use a deep Convolutional Neural Network (CNN) to extract features from the bounding box around the person in each time step. The output of the CNN is directly fed to an LSTM layer that captures temporal dynamics for each person. By adding a classification layer (i.e. softmax) on top of the first LSTM layer and using an end-to-end training of both CNN and the LSTM layer, we are able to learn discriminative image-based features for action recognition from individual human tracklets.

At each time step, the memory content of the first LSTM layer contains discriminative information describing the subject's action as well as past changes in his action. If the memory content is correctly collected over all people in the scene, it can be used to describe the group activity in the whole scene. 

Moreover, it can also be observed that direct image-based features extracted from the spatial domain around a person carries a discriminative signal for the ongoing activity. Therefore, a deep CNN model is used to extract complex features for each person in addition to the temporal features captured by the first LSTM layer. 

At this moment, the concatenation of the CNN features and the LSTM layer represent temporal features for a person. Various pooling strategies can be used to aggregate these features over all people in the scene at each time step. The output of the pooling layer forms our representation for the group activity. The second LSTM network, working on top of the temporal representation, is used to directly model the \textbf{temporal dynamics of group activity}. The LSTM layer of the second network is directly connected to a classification layer in order to detect group activity classes in a video sequence. 

Mathematically, the pooling layer can be expressed as the following:
 \begin{flalign}
  P_{tk} &= x_{tk} \oplus h_{tk} \\
  Z_{t} &= P_{t1} \diamond P_{t2} \ ... \diamond P_{tk}
\end{flalign}

In this equation, $h_{tk}$ corresponds to the first stage LSTM output, and $x_{tk}$ corresponds to the AlexNet fc7 feature, both obtained for the k\textsuperscript{th} person at time t. We concatenate these two features (represented by $\oplus$) to  obtain the temporal feature representation $P_{tk}$ for k\textsuperscript{th} person. We then construct the frame level feature representation $Z_{t}$ at time t by applying a max pooling operation (represented by $\diamond$) over the features of all the people. Finally, we feed the frame level representation to our second LSTM stage that operates similar to the person level LSTMs that we described in the previous subsection, and learn the group level dynamics. $Z_{t}$, passed through a fully connected layer, is given to the input of the second-stage LSTM layer.  The hidden state of the LSTM layer represented by $h_{t}^{group}$ carries temporal information for the whole group dynamics. $h_{t}^{group}$ is fed to a softmax classification layer to predict group activities. 
% \begin{flalign}
%  h^{fr}_{t} &= F(h^{fr}_{t - 1}, Z_{t})
% \end{flalign}
 
%Here $h^{fr}_{t}$ corresponds to the representation of t\textsuperscript{th} frame obtained by a transformation $F$ that is identical to the set of transformations explained in the previous subsection.

 %Next, we describe the implementation details that were used for constructing our model.

%The second stage model consists of a single LSTM layer that learns to predict frame level group activity labels. For doing this, we aggregate the person level features to generate frame level features. We try different aggregation techniques for generating the frame level features. Specifically, we consider max-pooling, average pooling, and sum pooling. We tried mimicing operation of cardinality kernel \cite{hajimirsadeghi2015visual} that yielded good performance in the past using our average pooling and sum pooling operation. However, as shown in thit turned out that they were not better than max pooling.

%  To propose people' locations for a foreground blob, we assume a fixed scale (w, h) for the person's bounding box (e.g. the moving camera almost has the same zoom). We use a simple brute force approach that tries every rectangle of scale (w, h) with step size (w/2, h/4) in the (right, bottom) directions assuming we start from the most top left location in the blob rectangle.

%-------------------------------------------------------------------------
\subsection{Implementation Details}
We trained our model in two steps. In the first step, the person-level CNN and the first LSTM layer are trained in an end-to-end fashion using a set of training data consisting of person tracklets annotated with action labels. 
%This model is very similar to the Recurrent Convolutional Neural Network (LRCN) proposed in \cite{donahue2014long} for action recognition\footnote{Implemented in \url{https://github.com/BVLC/caffe/pull/1873}}. 
We implement our model using Caffe~\cite{Caffe}. Similar to other approaches \cite{donahue2014long, DengZCLMRM15, venugopalan2014translating}, we initialize our CNN model with the pre-trained AlexNet network
%\footnote{Available at \url{https://github.com/BVLC/caffe/wiki/Model-Zoo}} 
and we fine-tune the whole network for the first LSTM layer. 9 timesteps and 3000 hidden nodes are used for the first LSTM layer and a softmax layer is deployed for the classification layer in this stage.

After training the first LSTM layer, we concatenate the fc7 layer of AlexNet and the LSTM layer for every person and pool over all people in a scene. The pooled features, which correspond to frame level features, are fed to the second LSTM network. This network consists of a 3000-node fully connected layer followed by a 9-timestep 500-node LSTM layer which is passed to a softmax layer trained to recognize group activity labels. 

For training all our models (that include both the baseline models and both the stages of the two-stage model), we follow the same training protocol. We use a fixed learning rate of 0.00001 and a momentum of 0.9. For tracking subjects in a scene, we used the tracker by Danelljan et al.~\cite{Danelljan14_tracker}, implemented in the Dlib library~\cite{dlib09}.

\section{Experiments}
\label{sec:experiments}
In this section, we evaluate our model by comparing our results with several baselines and previously published works on the Collective Activity Dataset~\cite{choi2009they} and our new volleyball dataset. 
First, we describe our baseline models. Then, we present our results on the Collective Activity Dataset followed by experiments on the volleyball dataset.

\subsection{Baselines}
The following baselines are considered in all our experiments:

\begin{enumerate}
\item \textbf{Image Classification: } %B1
This baseline is the basic AlexNet model fine-tuned for group activity recognition in a single frame.

\item \textbf{Person Classification:} % B4
In this baseline, the AlexNet CNN model is deployed on each person, fc7 features are pooled over all people, and are fed to a softmax classifier to recognize group activities in each single frame. %The whole network is fine-tuned from the pre-trained AlexNet model.

\item \textbf{Fine-tuned Person Classification:} %B4'
This baseline is similar to the previous baseline with one distinction. The AlexNet model on each player is fine-tuned to recognize person-level actions. Then, fc7 is pooled over all players to recognize group activities in a scene without any fine-tuning of the AlexNet model. The rational behind this baseline is to examine a scenario where person-level action annotations as well as group activity annotations are used in a deep learning model that does not model the temporal aspect of group activities. This is very similar to our two-stage model without the temporal modeling.

%that learning is done in two stages. First, action-level classifier is trained and then features representing players are used to train group-activities.

\item \textbf{Temporal Model with Image Features:} %B2
This baseline is a temporal extension of the first baseline. It examines the idea of feeding image level features directly to a LSTM model to recognize group activities. In this baseline, the AlexNet model is deployed on the whole image and resulting fc7 features are fed to a LSTM model. This baseline can be considered as a reimplementation of Donahue et al.~\cite{donahue2014long}.

\item \textbf{Temporal Model with Person Features:} %B3
This baseline is a temporal extension of the second baseline: fc7 features pooled over all people are fed to a LSTM model to recognize group activities.

\item \textbf{Two-stage Model without LSTM 1:} %B3
This baseline is a variant of our model, omitting the person-level temporal model (LSTM 1).  Instead, the person-level classification is done only with the fine-tuned person CNN.

\item \textbf{Two-stage Model without LSTM 2:} %B3
This baseline is a variant of our model, omitting the group-level temporal model (LSTM 2).  In other words, we do the final classification based on the outputs of the temporal models for individual person action labels, but without an additional group-level LSTM.

\end{enumerate}

\subsection{Experiments on the Collective Activity Dataset}
The Collective Activity Dataset \cite{choi2009they} has been widely used for evaluating group activity recognition approaches in the computer vision literature \cite{amer2014hirf, DengZCLMRM15, amer2012cost}. This dataset consists of 44 videos, eight person-level pose labels (not used in our work), five person level action labels, and five group-level activities. A scene is assigned a group activity label based on the majority of what people are doing. We follow the train/test split provided by \cite{hajimirsadeghi2015visual}. In this section, we present our results on this dataset.

\setlength{\tabcolsep}{1pt}
  \begin{table}[ht]
  \begin{center}
  \begin{tabular}{|l|c|}
  \hline
  Method & Accuracy\\
  \hline
  \hline
  B1-Image Classification                  & 63.0 \\
  B2-Person Classification                 & 61.8 \\ 
  B3-Fine-tuned Person Classification      & 66.3 \\ \hline
  B4-Temporal Model with Image Features    & 64.2 \\
  B5-Temporal Model with Person Features   & 62.2 \\ \hline
  B6-Two-stage Model without LSTM 1   & 70.1 \\
  B7-Two-stage Model without LSTM 2   & 76.8 \\ \hline
  \bf Two-stage Hierarchical Model	  & \bf 81.5	\\ \hline
  \end{tabular}
  \end{center}
  \caption{Comparison of our method with baseline methods on the Collective Activity Dataset.}
  \label{tab:acc_cad}
  \end{table}
  
  \setlength{\tabcolsep}{1pt}
  \begin{table}[ht]
  \begin{center}
  \begin{tabular}{|l|c|}
  \hline
  Method & Accuracy\\
  \hline
  \hline
  Contextual Model \cite{LanWYRM12}         &   79.1 \\ \hline
  Deep Structured Model \cite{DengZCLMRM15} &   80.6 \\ \hline
  \bf Our Two-stage Hierarchical Model & 81.5 \\ \hline
  Cardinality kernel \cite{hajimirsadeghi2015visual} &	\bf 83.4 \\ \hline
  \end{tabular}
  \end{center}
  \caption{Comparison of our method with previously published works on the Collective Activity Dataset.}
  \label{acc_cad_comp}
  \end{table}
  
  In Table~\ref{tab:acc_cad}, the classification results of our proposed architecture is compared with the baselines. As shown in the table, our two-stage LSTM model significantly outperforms the baseline models.
  An interesting comparison can be made between temporal and frame-based counterparts including B1 vs.\ B4,  B2 vs.\ B5 and B3 vs.\ our two-stage model. It is interesting to observe that adding temporal information using LSTMs improves the performance of these baselines.

  Table~\ref{acc_cad_comp} compares our method with state of the art methods for group activity recognition. The performance of our two-stage model is comparable to the state of the art methods. Note that only Deng et al.~\cite{DengZCLMRM15} is a previously published deep learning model. We postulate that there would be a significant improvement in the relative performance of our
  model if we had a larger dataset for recognizing group activities. In contrast, the cardinality kernel approach~\cite{hajimirsadeghi2015visual} outperformed our model. It should be noted that this approach works on hand crafted features fed to a model highly optimized for a cardinality problem (i.e.\ counting the number of actions in the scene) which is exactly the way group activities are defined in this dataset.
  
  \subsubsection{Discussion}
 
\begin{figure}[h]
  \includegraphics[width=1.0\linewidth]{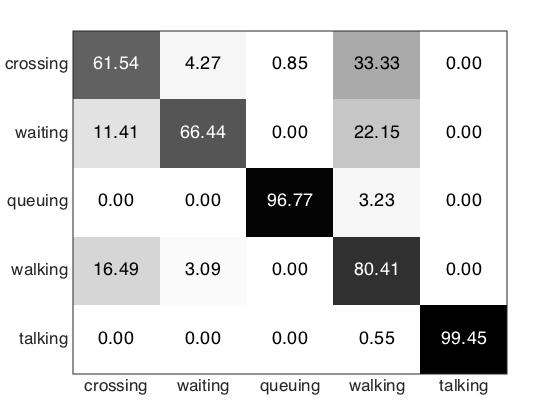}
    \caption{Confusion matrix for the Collective Activity Dataset obtained
    using our two-stage model.}
  \label{cad_cm}
  \end{figure}

  The confusion matrix obtained for the Collective Activity Dataset using our two-stage model is shown in Figure~\ref{cad_cm}. We observe that the model performs almost perfectly for the talking and queuing classes, and gets confused between crossing, waiting, and walking. Such behaviour is perhaps due to a lack of consideration of spatial relations between people in the group, which is shown to boost the performance of previous group activity recognition methods: e.g.\ crossing involves the walking action, but is confined in a path which people perform in orderly fashion. Therefore, our model that is designed only to learn the dynamic properties of group activities often gets confused with the walking action.
  
  It is clear that our two-stage model has improved performance with compared to baselines. The temporal information improves performance. Further, finding and describing the elements of a video (i.e.\ persons) provides benefits over utilizing frame level features.
  
 \subsection{Experiments on the Volleyball Dataset}

  In order to evaluate the performance of our model for team activity recognition on sport footage, we collected a new dataset based on publicly available YouTube volleyball videos. We annotated 1525 frames that were handpicked from 15 videos with seven player action labels and six team activity labels. We used frames from 2/3$^{rd}$ of the videos for training, and the remaining 1/3$^{rd}$ for testing. The list of action and activity labels and related statistics are tabulated in Tables \ref{scene_voll} and 
\ref{action_tab}.

\setlength{\tabcolsep}{1pt}
 \begin{table}[ht]
\begin{minipage}[b]{0.4\linewidth}
  \centering
  \begin{tabular}{|l|c|}
  \hline
  Group          & No. of \\
  Activity Class & Instances \\
  \hline
  \hline
   Right set			&	229	\\ \hline
   Right spike			&	187	\\ \hline
   Right pass           &   267 \\ \hline
   Left pass	    	&	304	\\ \hline
   Left spike 	    	&	246	\\ \hline
   Left set	            &	223 \\	\hline
  \end{tabular}
  \caption{Statistics of the group activity labels in the volleyball dataset.}
  \label{scene_voll}
  \end{minipage}
  \hspace{.05\linewidth}
  \begin{minipage}[b]{0.4\linewidth}
  \centering
  \begin{tabular}{|l|c|}
  \hline
  Action  & Average No. of\\
  Classes & Instance per Frame\\
  \hline
  \hline
   Waiting          &   0.30 \\ \hline
   Setting		    &	0.33 \\ \hline
   Digging          &   0.57 \\ \hline
   Falling          &   0.21 \\ \hline
   Spiking          &   0.28 \\ \hline
   Blocking         &   0.58 \\ \hline
   Others           &   9.22 \\ \hline
    
  \end{tabular}
   \caption{Statistics of the action labels in the volleyball dataset. \\}
  \label{action_tab}
  \end{minipage}
  \end{table}
  
  \begin{figure*}[ht]
\centering
\def\tabularxcolumn#1{m{#1}}
\begin{tabular}{cc}
\vspace{-0.2cm}
\subfloat[]{\includegraphics[width=6.5cm]{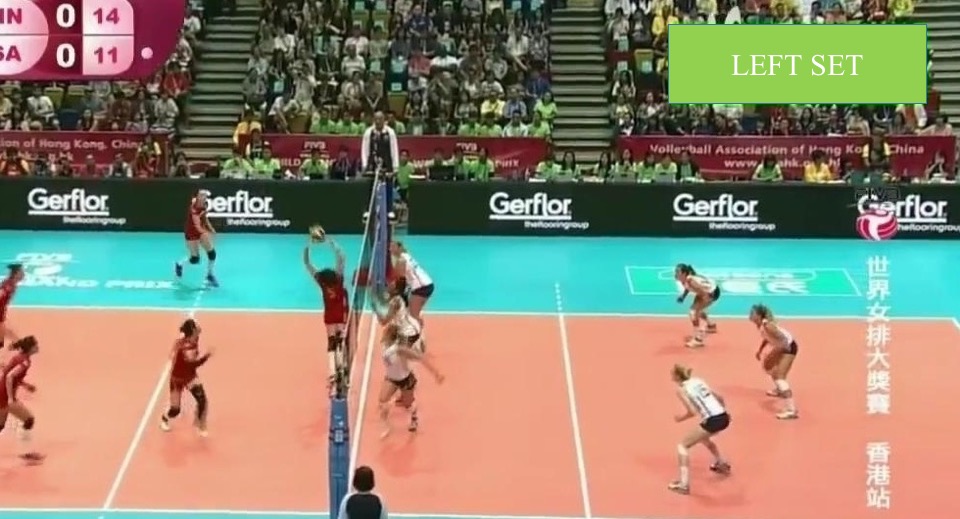}} 
   & \subfloat[]{\includegraphics[width=6.5cm]{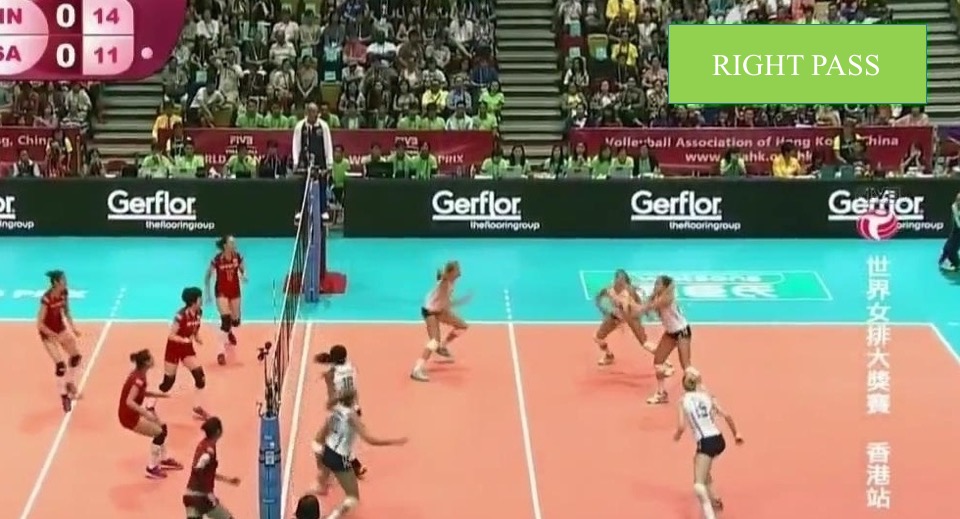}}\\ 
\vspace{-0.2cm}
\subfloat[]{\includegraphics[width=6.5cm]{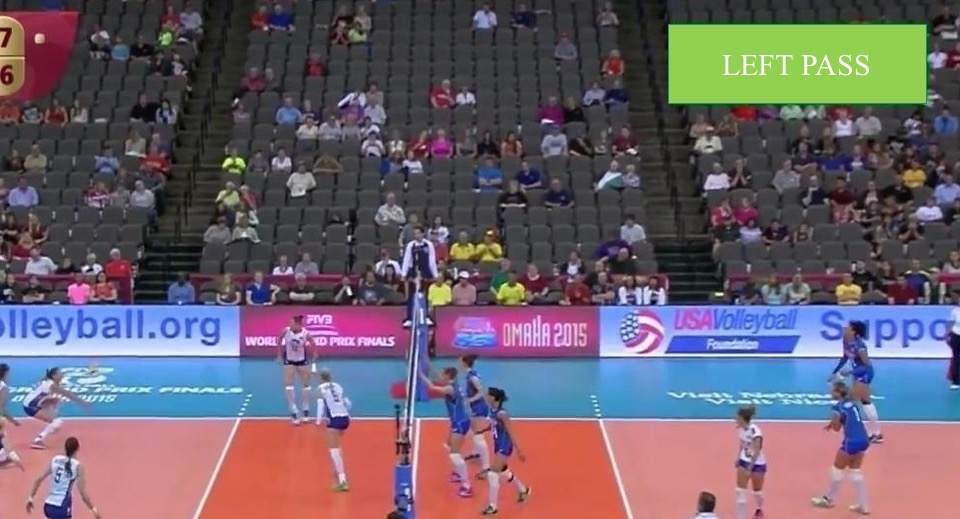}} 
   & \subfloat[]{\includegraphics[width=6.5cm]{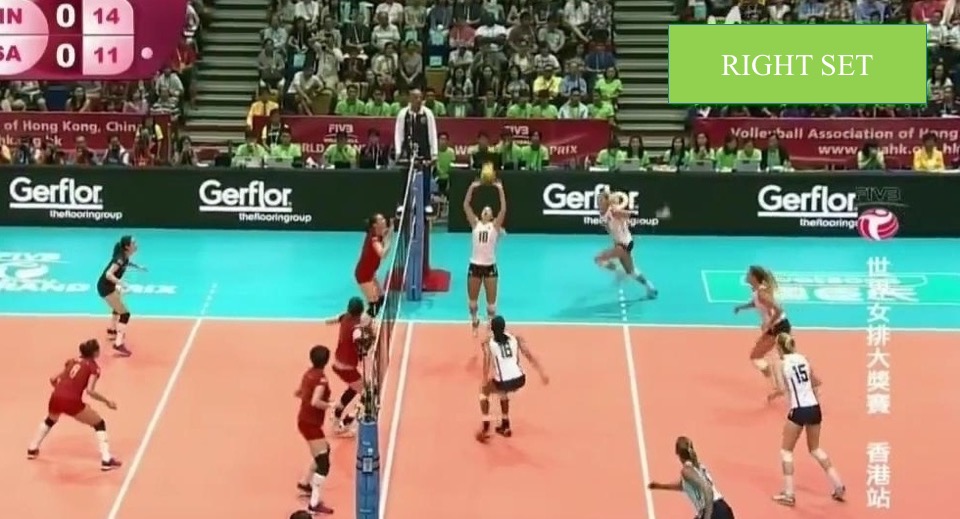}}\\
% Rebuttal - removing 2 images to save space for writing
% \vspace{-0.2cm}
% \subfloat[]{\includegraphics[width=6.5cm]{vis/50790.jpg}} 
%    & \subfloat[]{\includegraphics[width=6.5cm]{vis/50850.jpg}}\\
\vspace{-0.2cm}
\subfloat[]{\includegraphics[width=6.5cm]{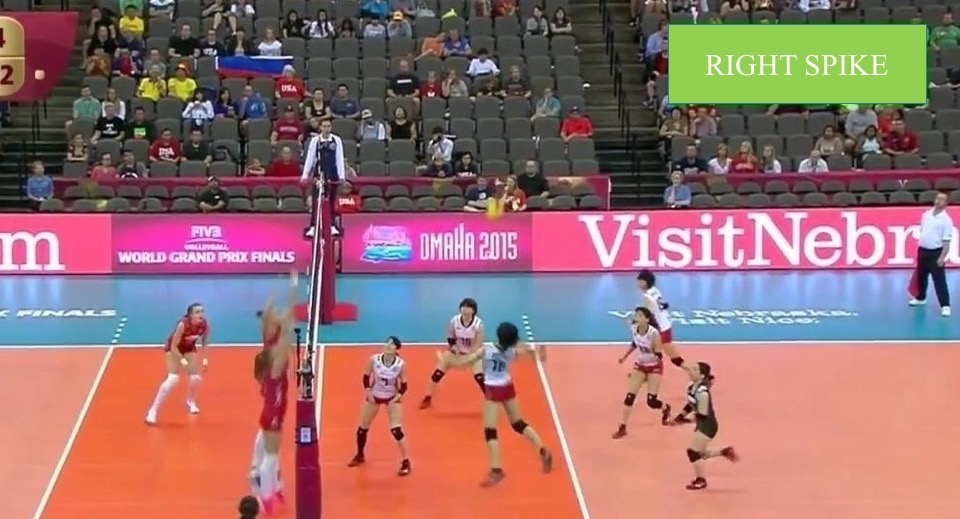}} 
   & \subfloat[]{\includegraphics[width=6.5cm]{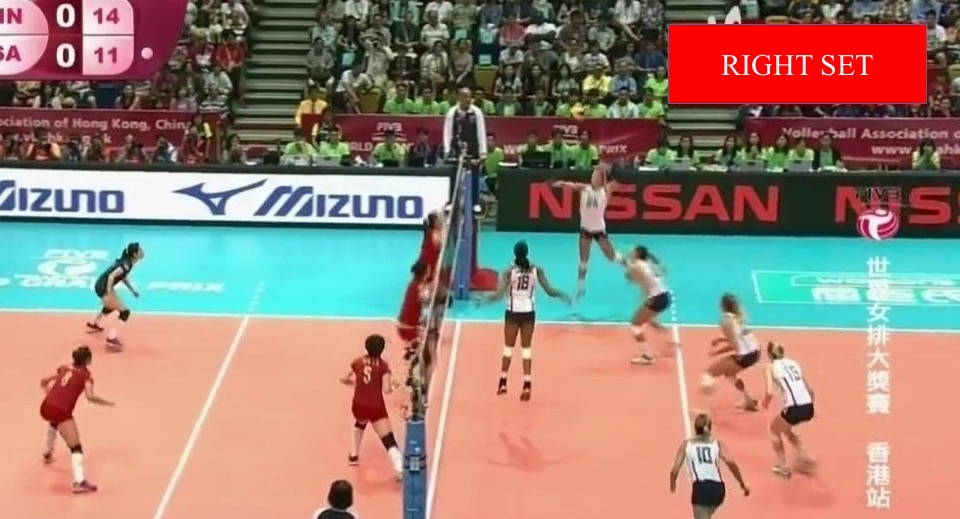}}\\
\vspace{-0.2cm}
\subfloat[]{\includegraphics[width=6.5cm]{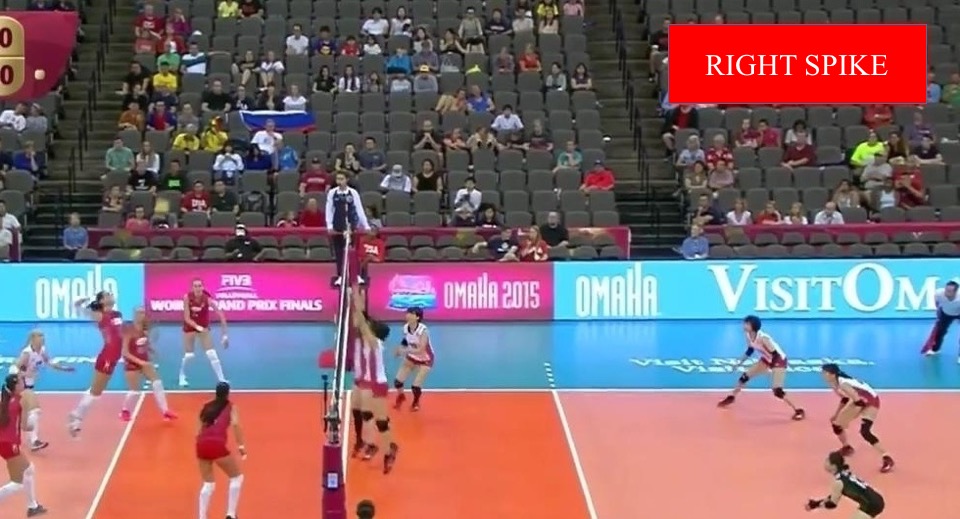}} 
   & \subfloat[]{\includegraphics[width=6.5cm]{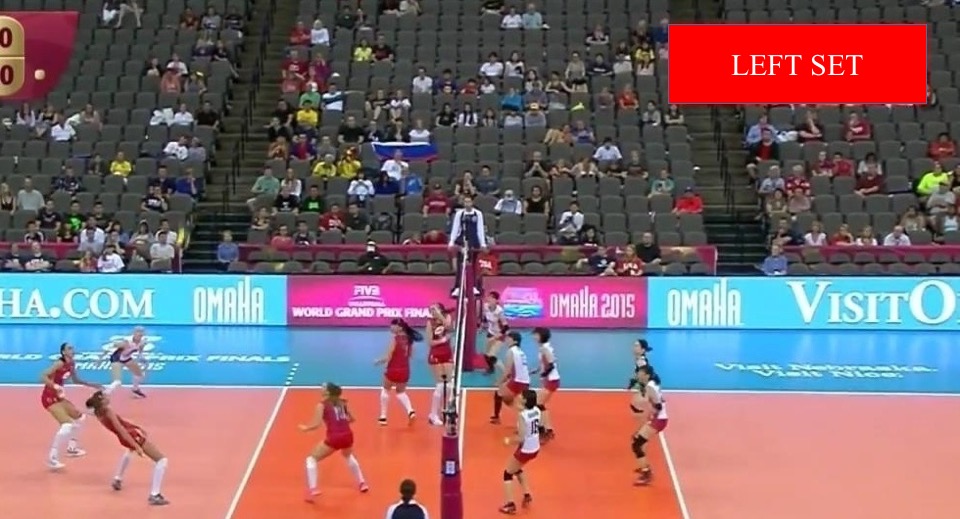}}\\
\end{tabular}

\caption{Visualizations of the generated scene labels using our model.  Green denotes correct classifications, red denotes incorrect. The incorrect ones correspond to the confusion between different actions in ambiguous cases (h and j examples), or in the left and right distinction (i example).}\label{fig:vis}
\end{figure*}

  From the tables, we observe that the group activity labels are relatively more balanced compared to the player action labels.
  This follows from the fact that we often have people present in static actions like standing compared to dynamic actions (setting, spiking, etc.). Therefore, our dataset presents a challenging team activity recognition task, where we have interesting actions that can directly determine the group activity occur rarely in our dataset. The dataset will be made publicly available to facilitate future comparisons \footnote{\url{https://github.com/mostafa-saad/deep-activity-rec}}.

  %In our experiment, we wished our first stage LSTM to learn representations pertaining to the important actions. For this purpose, we used six labels for the first stage instead of nine, by ignoring unimportant actions that include standing, moving and jumping. In the person level feature extraction stage, we allowed our model to overfit on persons who were performing these actions. This step was essential for us to focus on key actions for recognizing group activities.
  
  In Table \ref{tab:acc_vol}, the classification performance of our proposed model is compared against the baselines. Similar to the performance in the Collective Activity Dataset, our two-stage LSTM model outperforms the baseline models. However, compared to the baselines, the performance gain using our model is more modest. This is likely because we can infer group activity in volleyball by using just a few frames. Therefore, in the volleyball dataset, our baseline B1 is closer to the actual model's performance, compared to the Collective Activity Dataset. Moreover, explicitly modeling people is necessary for obtaining better performance in this dataset, since the background is rapidly changing due to a fast moving camera, and therefore it corrupts the temporal dynamics of the foreground. This could be verified from the performance of our baseline model B4, which is a temporal model that does not consider people explicitly, showing inferior performance compared to the baseline B1, which is a non-temporal image classification style model. On the other hand, baseline model B5, which is a temporal model that explicitly considers people, performs comparably to the image classification baseline, in spite of the problems that arise due to tracking and motion artifacts. 
  
  \setlength{\tabcolsep}{1pt}
  \begin{table}[ht]
  \begin{center}
  \begin{tabular}{|l|c|}
  \hline
  Method & Accuracy\\
  \hline
  \hline
  B1-Image Classification                  & 46.7 \\
  B2-Person Classification                 & 33.1 \\ 
  B3-Fine-tuned Person Classification      & 35.2 \\ \hline
  B4-Temporal Model with Image Features    & 37.4 \\
  B5-Temporal Model with Person Features   & 45.9 \\ \hline
  B6-Our Two-stage Model without LSTM 1   & 48.8 \\
  B7-Our Two-stage Model without LSTM 2   & 49.7 \\ \hline
  \bf Our Two-stage Hierarchical Model	   & \bf 51.1	\\ \hline
  \end{tabular}
  \end{center}
  \caption{Comparison of the team activity recognition performance of baselines against our model evaluated on the volleyball dataset.}
  \label{tab:acc_vol}
  \end{table}
  
  In both datasets, an observation from the tables is that while both LSTMs contribute to overall classification performance, having the
first layer LSTM (B7 baseline) is relatively more critical to the performance of the system, compared to the second layer LSTM (B6 baseline).

  All the reported experiments use max-pooling as mentioned above. However, we also tried both sum and average pooling, but their performance was consistently lower compared to their max-pooling counterpart.

  %In the volleyball dataset, we consider 2 baselines versus our structured model. In the first baseline experiment, we learn pure scene classification: categorize a frame to one of the group activities with no temporal information considered. To train such network, we fine-tuned AlexNet with training examples of (frame, activity label) pairs. \\

  %In the second baseline, we do also scene classification but though a temporal window over K tracklets. Specifcally, we detect K players in the target testing frame and track them through window of W frames centered at that frame (hence capture information before and after the event). A W-reshaped frames for the K bounding boxes are built as in Figure~\ref{fig_detect_reshape_baseline2}. To train a spatio-temporal model for that, we feed the reshaped frames to a network composed of AlexNet followed by LSTM layer. 

%\begin{figure}
%    \centering
%    \subfloat[Detections]{{\includegraphics[height=2.5cm]{images/00026591.jpg} }}
%    \hspace{0.7pt}   
%    \subfloat[Reshaped Image]{{\includegraphics[height=2.5cm,width=3cm]{images/00026591_reshape.jpg} }}
%    \caption{Baseline 2: players reshaping. In the left image, we detect the K people and then reshape them as in the right image. As classifier is not accuracte, K is slightly bigger than the actual number of players to avoid missing them.}
%    \label{fig_detect_reshape_baseline2}
%\end{figure}

\begin{figure}[h]
  \includegraphics[width=1.0\linewidth]{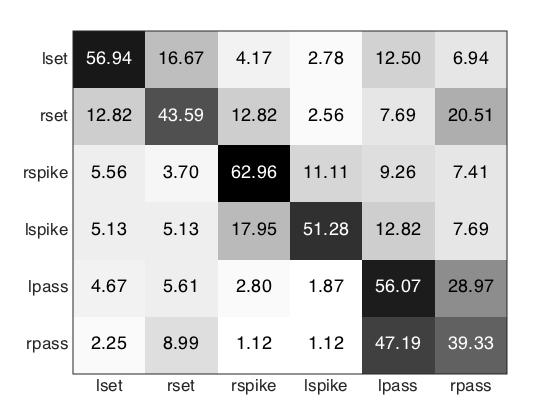}
    \caption{Confusion matrix for the volleyball dataset obtained
    using our two-stage hierarchical model.}
  \label{cm_voll}
  \end{figure}

\subsubsection{Discussion}

  Figures~\ref{fig:vis} and \ref{cm_voll} show  visualizations of our detected activities and the confusion matrix obtained for the volleyball dataset using our two-stage model. From the confusion matrix, we observe that our model generates consistently accurate high level action labels. Nevertheless, our model has some confusion between set and pass activities, as these activities often may look similar.
  
  %problem with identifying the side that is performing that action. {\color{red} This is evident from high values observed for the wrong side as shown in the confusion matrix.}

%  \begin{figure}[t]
%  \begin{center}
%  \includegraphics[width=0.8\linewidth]{images/00026591.jpg}
%  \end{center}
%    \caption{{\bf Visualization of our results on the volleyball dataset.} .... show some images, boxes in it, show the LSTM1 judging of actions and overall activity.}
%  \label{fig_compare_sucess_fail}
%  \end{figure}

%-------------------------------------------------------------------------
\section{Conclusion}
\label{sec:conclusion}
In this paper, we presented a novel deep structured architecture to deal with the group activity recognition problem. Through a two-stage process, we learn a temporal representation of person-level actions and combine the representation of individual people to recognize the group activity. We also created a new volleyball dataset to train and test our model, and also
evaluated our model on the Collective Activity Dataset. Results show that our architecture can improve upon baseline methods lacking hierarchical consideration of individual and group activities using deep learning.

%In the future work, we would like to try complex fusion styles and compare their different performance. e.g. We could improve the performance in volleyball dataset by a model that is explicitly designed to identify the side that performs the action. In addition, we plan to gather a bigger volleyball dataset and evaluate our models against it.

\section*{Acknowledgements}

This work was supported by grants from NSERC and Disney Research.

{\small
\bibliographystyle{ieee}
\bibliography{egpaper_for_review}
}

\end{document}